# Soft Autoencoder and Its Wavelet Adaptation Interpretation

Fenglei Fan, *Student Member, IEEE,* Mengzhou Li, Yueyang Teng, Ge Wang*, *Fellow, IEEE*

***Abstract*—Recently, deep learning becomes the main focus of machine learning research and has greatly impacted many important fields. However, deep learning is criticized for lack of interpretability. As a successful unsupervised model in deep learning, the autoencoder embraces a wide spectrum of applications, yet it suffers from the model opaqueness as well. In this paper, we propose a new type of convolutional autoencoders, termed as Soft Autoencoder (Soft-AE), in which the activation functions of encoding layers are implemented with adaptable soft-thresholding units while decoding layers are realized with linear units. Consequently, Soft-AE can be naturally interpreted as a learned cascaded wavelet shrinkage system. Our denoising experiments demonstrate that Soft-AE not only is interpretable but also offers a competitive performance relative to its counterparts. Furthermore, we propose a generalized linear unit (GenLU) to make an autoencoder more adaptive in nonlinearly filtering images and data, such as denoising and deblurring.**

***Index Terms*—Deep learning, Interpretability, Convolutional Autoencoder, Activation functions.**

## I. Introduction

Deep learning has over recent years made huge strides in many important fields [1-4]. As a successful unsupervised learning model, autoencoders such as denoising autoencoder [5], contractive autoencoder [6], k-sparse autoencoder [7], variational autoencoder (VAE) [8], and convolutional autoencoder [9] play significant roles in feature extraction, denoising, dimension reduction, generative tasks, and so on. However, akin to other deep learning models, an autoencoder suffers from lack of interpretability. Currently, it is still difficult to understand the mechanism of the autoencoder, let alone to have any governing guideline for the optimal design of an autoencoder in a task-specific fashion. As a result, only empirical exploration serves as the basis for auto-encoder prototyping.

Due to the importance of interpretability, considerable efforts have been made in explaining the mechanism of deep learning such that more trust can be placed on the autoencoder to push the boundary of its applications. The existing methods that explain neural networks can be categorized into four classes [10]: hidden neuron analysis [11], model mimicking methods [12-13], localized interpretation methods [14-15], and physics/engineering methods [16]. The hidden neuron analysis methods interpret a neural network by visualizing or dissecting the features extracted by hidden neurons. The model mimicking methods build explainable models that deliver the performance as closely as possible to that of the "black-box" models. Given trained neural networks, the local interpretation methods investigate the importance of inputs by perturbing the input and analyzing changes in the resultant output. Lastly, the physics/engineering methods find significant connections between deep networks and advanced physical or engineering systems to reveal the mechanisms of neural networks. Note that such a classification is qualitative and imprecise, some methods can be put into multiple classes from different perspectives. For example, our fuzzy logic interpretation method [17] analyzes the spectrum of every quadratic neuron and can be viewed as either hidden neuron analysis or engineering modeling.

In this manuscript, as shown in Figure 1, we propose an interpretable convolutional autoencoder, termed as the soft autoencoder (Soft-AE), in which the activation functions in the encoding layers are implemented with adaptable soft-thresholding units [18] $\eta_{b<0}(x) = sgn(x)\max\{|x| + b, 0\}$, where $b$ is a threshold, $sgn(\cdot)$ is the sign function, and the decoding layers are equipped with linear units. With such a configuration, Soft-AE performs a network-based wavelet transform embedded with soft thresholding shrinkage operations. Hence, a deep Soft-AE system can be naturally interpreted as a learned deep and cascaded wavelet shrinkage system. The convolutional autoencoder is a special type of autoencoders, which is intrinsically more appropriate for image denoising and some other tasks compared to the counterparts in the form of multi-layer perceptrons (MLP). When dealing with image formation and analysis, a fully connected autoencoder is unrealistic due to the memory requirement and unnecessary redundancy in the space of parameters. In contrast, the convolutional autoencoder incorporates convolution/deconvolution operations in its encoding and decoding processes, thereby reducing network redundancy and computational overhead, permitting multi-resolution analysis in a nonlinear fashion. Furthermore, we theoretically investigate the resolution enhancing property of $\eta_{b>0}(x) = sgn(x)\max\{|x| + b, 0\}$, in contrast to the soft thresholding unit $\eta_{b<0}(x)$. Then, we present a generalized lineal unit (GenLU) as novel activation functions to enhance the autoencoder for more image processing tasks from denoising to deblurring.

The contributions of our work are three folds: First, in the context of convolutional auto-encoding, we make an effort to link deep learning to contemporary signal processing [19]. In this aspect, we bridge classical wavelet analysis and deep convolutional auto-encoding by modifying activations in a convolutional autoencoder such that the wavelet shrinkage scheme is absorbed inside the autoencoder. Second, we employ

This work was partially support by IBM AI Horizon Scholarship, NIH/NCI under award numbers R01CA233888 and R01CA237267, and NIH/NIBIB under award number R01EB026646.

Fenglei Fan (fanf2@rpi.edu), Mengzhou Li (lim23@rpi.edu) and Ge Wang* (E-mail: wangg6@rpi.edu) are with the Department of Biomedical Engineering, Rensselaer Polytechnic Institute, Troy, NY, USA, 12180. Yueyang Teng is with Sino-Dutch Biomedical and Information Engineering School, Northeastern University, Shenyang, China, 110169. Asterisk indicates the corresponding author.

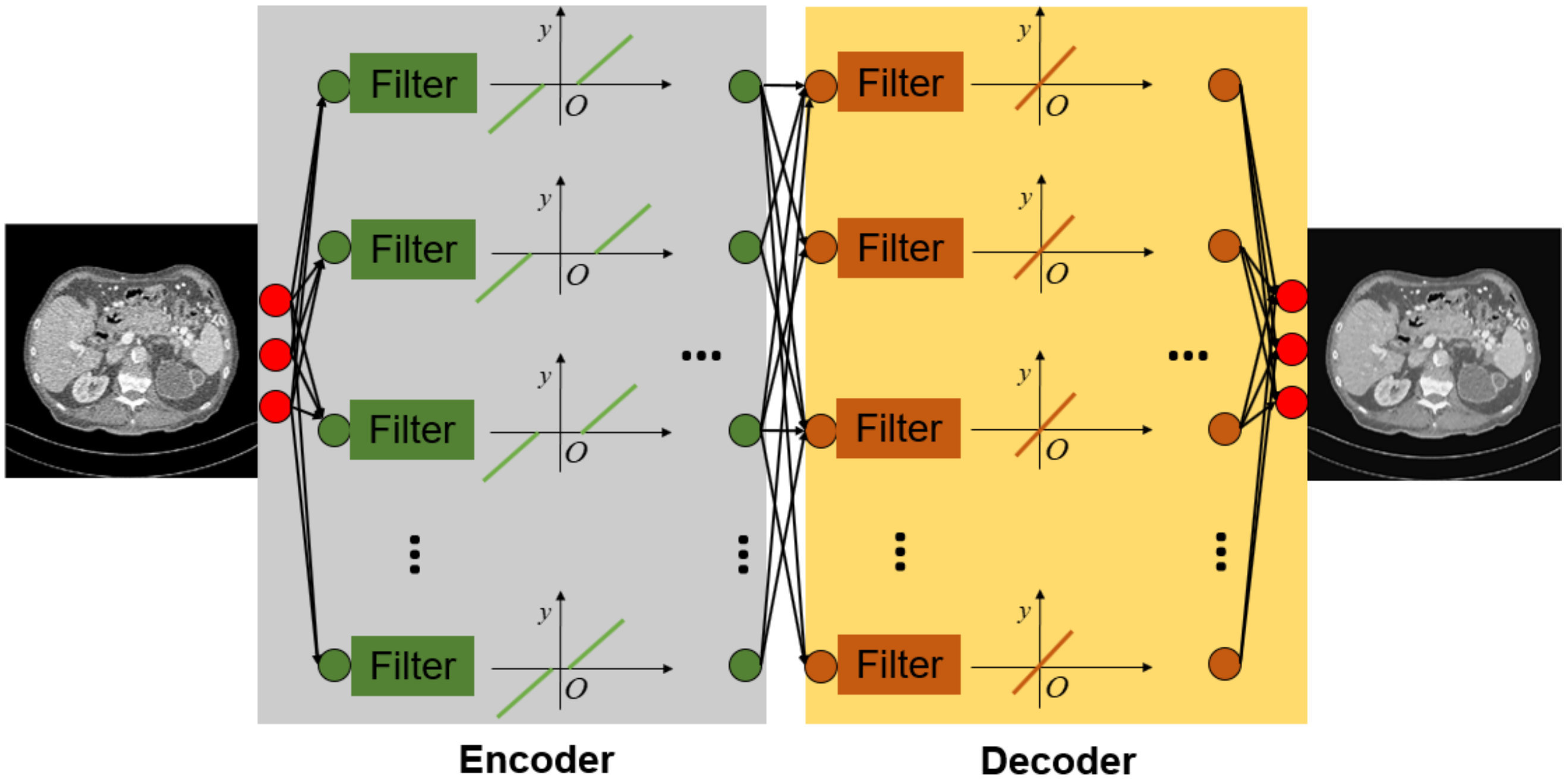

Figure 1. Soft-autoencoder interpreted as a wavelet shrinkage system after activation functions are appropriately made for image denoising.

soft thresholding units, which is a new way to look at an activation function. In the framework of Soft-AE, wavelets and thresholds for soft-thresholding are learned in the training stage from big data. Such a character enables Soft-AE to embrace big-data-empowered capability and robustness in contrast to traditional wavelet analysis since most comprehensive knowledge is contained in the big data. Our experiments demonstrate that Soft-AE performs competitively on various benchmarks. Third, we further propose a novel activation function called "generalized linear unit (GenLU)" for diverse tasks.

To put our contributions in perspective, let us review the relevant studies as follows. (1) The activation unit ReLU is the most popular nonlinear activation function in deep learning because it is able to prevent gradients from vanishing or exploding. However, ReLU also arguably tends to block the circulation of information. The concatenated ReLU [20], Max-Min Networks [21], ON/OFF ReLU [22], Concatenated-ReLU [23], and Leaky-ReLU [24] dedicated to taking more information. (2) Efforts were recently made to interpret autoencoders. Zhao *et al.* [25] proposed the stacked what-where autoencoder (SWWAE) to reduce information loss. In SWWAE, the location information is incorporated for signal recovery/reconstruction. Yang *et al.* [26] utilized invertible functions to build an autoencoder which facilitates interpretability. Adel *et al.* [27] used the generative model such as the normalized flow method to transform the hidden representations of an autoencoder to the disentangled representations, where the degree of disentanglement was computed based on testing digit samples containing prior information (thickness, skewness, etc.). Yu and Principe (2019) [28] applied information bottleneck theory to describe the information flow pertaining to the mutual information states of symmetric layers in a stacked autoencoder as

$$I(X;X') \geq I(T_1;T_1') \geq \cdots \geq I(T_K;T_K'),$$

where $X$ and $X'$ are the input and the yield of the autoencoder respectively, $T_i$ and $T_i'$ are the outputs of the $i^{th}$ symmetric layers in the encoder and decoder respectively. Higgins *et al.* [29] developed the β-VAE to enforce the disentanglement regularization, which relies on the KL distance between the distribution of latent factors and their posterior distribution. Later, Hsu *et al.* [30] established an interpretable VAE for sequential data by imposing sequence-dependent and sequence-independent priors to different groups of latent variables. (3) The applications of wavelets in neural networks were investigated already [31-35]. Particularly, the scattered wavelet network was developed to iteratively collect coefficients of a scattered wavelet transform at different scales. Note that there are distinctions between a scattered wavelet network and a deep convolutional network. Most noticeably, the wavelet representation is derived from the output of all layers instead of just the final layers, and the filters are not learned from the data but from the predefined wavelet filters. Also note that the downstream applications are based on the coefficients of those wavelet transforms. In contrast, our soft autoencoder is designed for the denoising task, and the filters are learned from data.

The study in Ye *et al.* [35] is most relevant to our work, in which the convolutional framelet theory with a low-rank Hankel matrix was leveraged to represent signals by their local and non-local bases, suggesting an encoding-decoding structure that promises a perfect signal reconstruction. Albeit providing a linearized interpretation, there are several aspects that can be enhanced. As mentioned in Remark 3 in [35], the non-local basis is a general pooling/un-pooling operator, however, the pooling reduces the dimension of data, un-natural to the representation framework. In addition, to tackle with the nonlinearity from ReLU, the authors combined two "opposite" ReLUs to transform the nonlinearity into the linearity so that a perfect recovery conditions can be argued. Although this trick is sound, it potentially hurts the power of deep learning because it counteracts the nonlinearity that is commonly accepted as a key ingredient of deep learning. In contrast, our model is analogous to a wavelet shrinkage system, where pooling and un-pooling operations are not needed to keep structural consistency. Therefore, our interpretation has no need to explain pooling and un-pooling. Furthermore, our model favorably accommodates the nonlinearity as the critical characteristic of the framework in the form of soft thresholding units.

## II. SOFT-AE AND GENERALIZED LINEAR UNIT

For completeness, let us first introduce relevant preliminaries as well as the wavelet shrinkage algorithm. Then we present the design of Soft-AE and shed the light on the conditions that traverse the gulf between a Soft-AE and a wavelet shrinkage system. Next, we propose a generalized linear unit (GenLU).

### *A. Wavelet Shrinkage System and Soft-AE*

#### *A. 1. Wavelet shrinkage system*

**Soft-thresholding:** Soft-thresholding [18] is an important tool in signal processing due to the effectiveness of wavelet shrinkage methods. Usually the soft-thresholding is superior to the hard-thresholding for two reasons. First, the theoretical analysis suggests that the soft-thresholding operation has a good property in terms of smoothness [18]; see the **Theorem** in Section II. B. 1. Second, the soft-thresholding gives continuous results, while the hard-thresholding is discontinuous and produces abrupt artifacts in the denoised images. Given an input, the soft thresholding unit will produce an output:

$$\eta_{b<0}(x) = sgn(x)\max\{|x| + b, 0\}, \tag{1}$$

where the threshold $b$ is empirically pre-determined in traditional domain, and $sgn(\cdot)$ is the sign function. In Soft-AE, $b$ will be advantageously learned in the training process from a training dataset.

**Wavelet transform:** The wavelet transform of $f(x)$ in terms of a wavelet $\Psi(x)$ is defined as follows:

$$[W_\Psi(f)](a, c) = \int_{-\infty}^{+\infty} \Psi\left(\frac{x-a}{c}\right) f(x)\, dx, \tag{2}$$

where $\Psi$ is a pre-determined wavelet. Common wavelets are Morlets, Daubechies wavelets, and so on. $[W_\Psi(f)](a, c)$ is called wavelet coefficients. For a specific resolution, the wavelet transform is equivalent to a convolution with a corresponding wavelet kernel at a specific scale. Therefore, in the following, we use wavelet transformation and convolution interchangeably.

**Wavelet Shrinkage Denoising:** Donoho [18] proposed the *VisuShrink* algorithm that uses the soft thresholding operation and enjoys optimal denoising properties. However, the *VisuShrink* algorithm tends to over-smoothen results. The *SureShrink* [36] combines a universal constant and a SURE threshold, derived for minimizing Stein's unbiased risk estimator. *BayesShrink* includes adaptive data-driven thresholds [37], which are set differently in sub-bands through Bayes estimation assuming that the wavelet coefficients in each sub-band are with the generalized Gaussian distribution. *SmoothShrink* reduces speckle noise by applying a directional smoothing function based convolutional kernel on the wavelet coefficients [38].

Basically, the *VisuShrink* algorithm consists of the following three steps in the pseudo-code below: (a) perform the wavelet transform to derive wavelet coefficients; (b) apply an element-wise soft-thresholding operation to the wavelet coefficients; (c) perform the inverse wavelet transform. Mathematically, suppose that we have the following additive noise model: $Y(t) = S(t) + N(t)$, where $Y(t)$ and $S(t)$ are measurement and the authentic signal respectively. Then, the above three steps will correspond to the following three formulas: $\hat{Y} = \mathrm{W}(Y)$ ; $\mathrm{Z} = \eta_{-\sigma_N\sqrt{2logn}}(\hat{Y})$, where $\sigma_N^2$ is the noise variance and $n$ is the number of pixels; $\hat{S} = W^{-1}(Z)$.

**_VisuShrink_ Algorithm**

**Input**: $Y(t) = S(t) + N(t)$, wavelet $\psi$
1: Wavelet transform by $\psi$: $\hat{Y} = W_\psi(Y)$
2: Soft thresholding: $\mathrm{Z} = \eta_{-\sigma_N\sqrt{2logn}}(\hat{Y})$.
3: Inverse wavelet transform by $\psi^{-1}$: $\hat{S} = W_\psi{}^{-1}(Z)$
**Output**: $\hat{S}(t)$

Here we heuristically illustrate why a soft thresholding unit works so well. As shown in Figure 2, the wavelet coefficients of a corrupted signal are full of glitches with small amplitudes over the whole spectrum. Evidently, linear estimators are not adequate to remove noise from wavelet coefficients, because noise is uneven and everywhere. When soft thresholding is applied in the wavelet domain, we have $|\mathrm{Z}| \le |W_\psi(\mathrm{S(t)})|$. Then, in the signal domain after the inverse wavelet transform, $\|\hat{S}\|_{B^s_{p,q}} \le C\|S\|_{B^s_{p,q}}$, where C is a constant and $\|\cdot\|_{B^s_{p,q}}$ represents the Besov norm. Suppose that $S(t)$ is a zero function, $\hat{S}(t)$ will also be a zero function, which means that the *VisuShrink* can obtain a smooth recovery at least in such an extreme case. On the contrary, some other estimators such as the hard-thresholding estimator exhibit annoying bumps even when reconstructing very smooth functions.

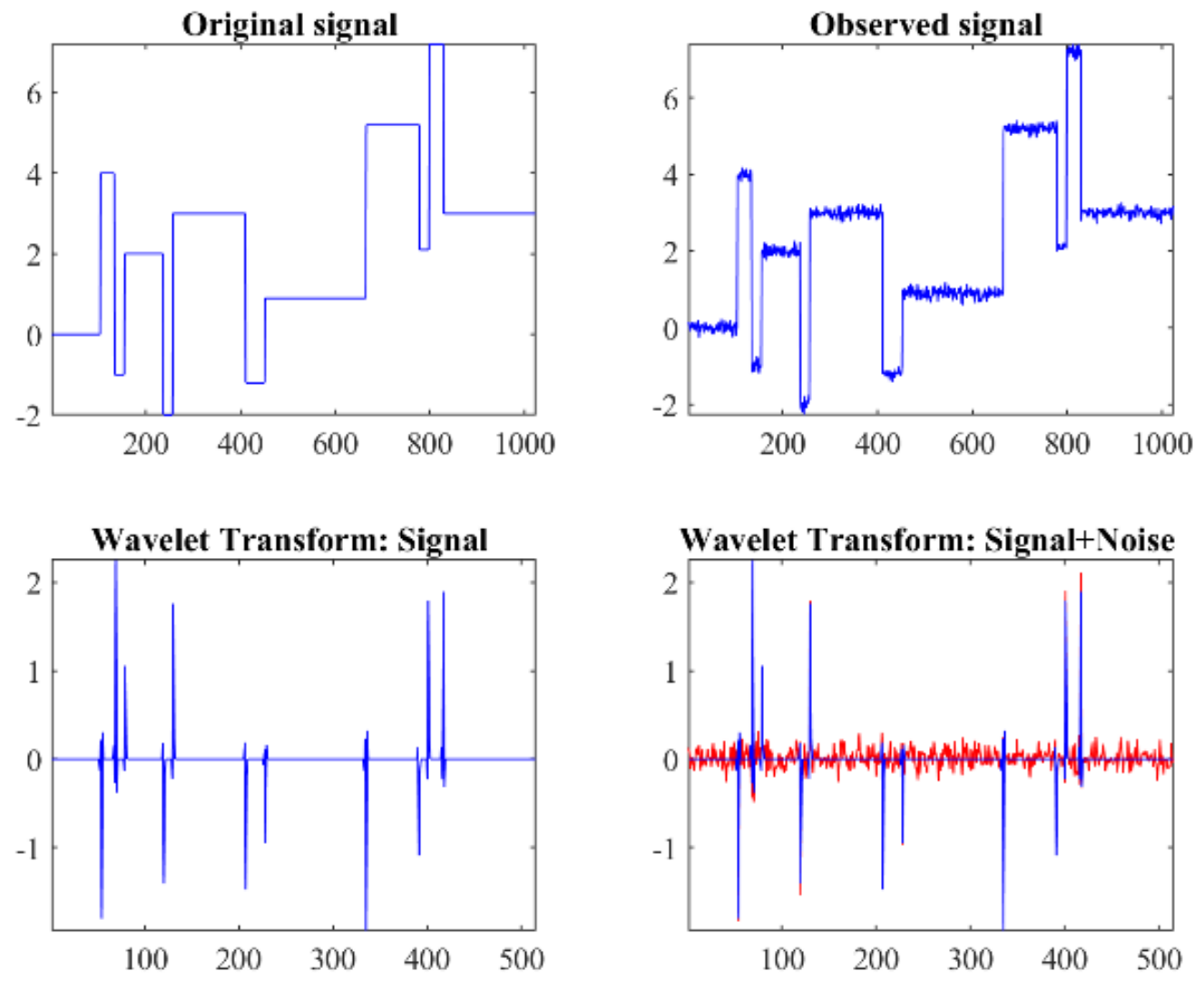

Figure 2. Soft thresholding in the wavelet domain.

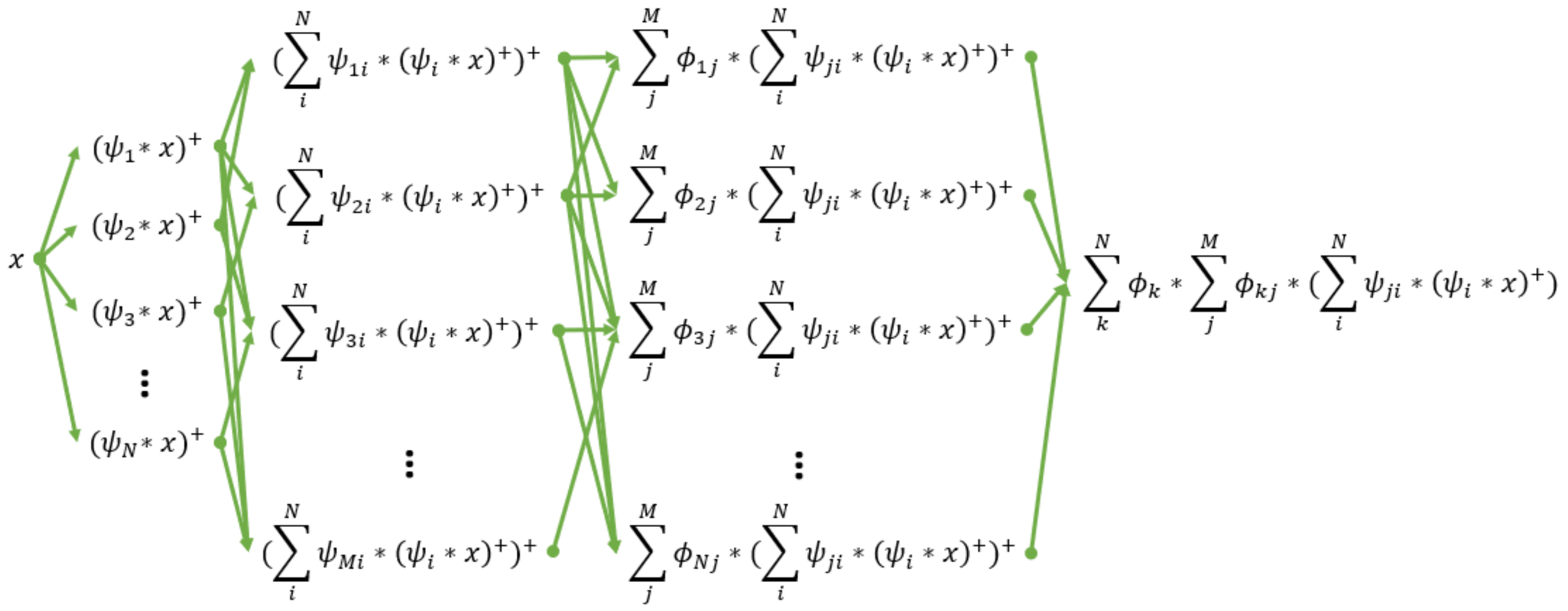

Figure 3. Overall computational process of Soft-AE through encoding and decoding operations. $(\cdot)^+$ represents the soft thresholding operation.

### *A. 2. Soft-AE*

Inspired by the success of the wavelet shrinkage system, we propose a novel type of convolutional autoencoder that deploys soft thresholding units as activation functions in the encoding layers and liner functions as activation functions in the decoding layers. In this regard, we facilitate interpretability and model adaptivity simultaneously for convolutional neural networks, turning a black-box convolutional autoencoder into an interpretable soft autoencoder (Soft-AE). In other words, the conventional three-step wavelet shrinkage system is a special case of Soft-AE, and **a Soft-AE is nothing but a learned cascade wavelet shrinkage system**. In Soft-AE, the discrete wavelet transformation and soft-thresholding operations are sequentially conducted in the encoding layers, and then decoding layers recover a desirable signal accordingly.

To put our scheme in perspective, let us do a general analysis and explain the relationship between Soft-AE and the wavelet shrinkage system. Let us start from a two-convolutional-layer Soft-AE and suppose that there are $N$ convolutional filters in each layer, denoted as $\psi_i$ (encoding layer) and $\phi_i$ (decoding layer). We use $*$ to represent convolution and superscript $+$ to represent soft-thresholding operation. Given the input $x$ of a finite length, the expression for the yield of a two-convolutional-layer Soft-AE can be expressed as

$$\sum_i^N \phi_i * (\psi_i * x)^+, \quad (5)$$

where $(\cdot)^+$ represents the soft thresholding operation. When the functions $\psi_i, \phi_i$ fulfill that

$$\phi_i = \frac{\psi_i^{-1}}{N} \quad or \quad \psi_i = \frac{\phi_i^{-1}}{N}, \quad (6)$$

where $(\cdot)^{-1}$ represents the inverse transform, Soft-AE with two convolutional layers makes a perfect match with the wavelet shrinkage system when $\psi_i$ is the inverse of $\phi_i$. Please note that Eq. (6) holds for common wavelets such as Morlets and Daubechies wavelets.

More generally, let us consider a four-convolutional-layer Soft-AE. Without loss of generality, we assume that there are $N$ filters in the first encoding layer and $M * N$ filters in the second encoding layer. The convolutional filters in the encoding layers are denoted as $\psi_i, i = 1,2,\dots,N$ and $\psi_{ij}, i = 1,2,\dots,M; j = 1,2,\dots N$ respectively. In symmetry, the two decoding layers have $N * M$ and $N$ filters respectively. We denote the deconvolutional filters in the decoding layers as $\phi_{ij}, i = 1,2,\dots,N; j = 1,2,\dots M$ and $\phi_i, i = 1,2,\dots,N$. Figure 3 illustrates the computational process of Soft-AE with four convolutional layers. The final output is

$$\sum_k^N \phi_k * [\sum_j^M \phi_{kj} * [\sum_i^N \psi_{ji} * (\psi_i * x)^+]^+], \quad (7)$$

where we can apply the property of the soft thresholding:

$$(h + g)^+ = h^+ + g^+, \quad (8)$$

which holds approximately when the magnitude of the threshold is small, and Eq. (7) reduces into

$$\sum_k^N \phi_k * [\sum_j^M \phi_{kj} * \sum_i^N [\psi_{ji} * (\psi_i * x)^+]^+]. \quad (9)$$

Suppose $\mathbf{\Psi}$ is the $M \times N$ matrix with $\psi_{ji}$ at its row $j$ and column $i$, while $\mathbf{\Phi}$ is the $N \times M$ matrix with $\phi_{kj}$ at its row $k$ and column $j$. Using the associative laws of convolution operation, we can further simplify Eq. (9) into the matrix form:

$$[\phi_1,..,\phi_N] \otimes \mathbf{\Phi} \otimes \mathbf{\Psi} \otimes [(\psi_1 * x)^+,..,(\psi_N * x)^+]^T, (10)$$

where $(A \otimes B)_{ij} = \sum_k A_{ik} * B_{kj}$, which is analogous to the matrix product but the involved elements here are functions, and the convolutional operation is performed between the elements. Therefore, for Soft-AE to realize wavelet shrinkage, the following conditions should be met:

$$\begin{cases} \mathbf{\Phi} \otimes \mathbf{\Psi} = \mathrm{diag}(\lambda_1, \lambda_2, \dots, \lambda_N)\delta \\ \phi_k = \dfrac{\psi_k^{-1}}{|\sum_k^N \lambda_k|} \quad or \ \psi_k = \dfrac{\phi_k^{-1}}{|\sum_k^N \lambda_k|}, k = 1,2,\dots,N \end{cases}, \quad (11)$$

where $\delta$ is the Dirac function, and $\sum_k^N \lambda_k$ is supposed to be non-zero that can be made by the selection of $\mathbf{\Phi} \otimes \mathbf{\Psi}$. The existence of $\mathbf{\Phi}$ and $\mathbf{\Psi}$ that fulfills Eq. (11) is natural, one trivial situation is that diagonal elements of $\boldsymbol{\Phi}$ and $\boldsymbol{\Psi}$ are mutually inverse to each other, and the rest elements are zero.

**Remark 1:** Our derivation is in the framework of Soft-AE, we offer the mapping between Soft-AE and a wavelet shrinkage system under the conditions that enable the Soft-AE to realize a wavelet shrinkage system. Please note that these conditions can be extended to deeper versions of Soft-AE through similar steps. The approximation in Eq. (8) we made on the soft thresholding is reasonable, as instantiated in Figure 2. When the noise intensity is small, the threshold value to be applied is small as well, which renders the soft thresholding unit close to a linear unit. Thus, the soft-thresholding operation to the addition of two signals can be decomposed into the addition of the soft-thresholding operations to each signal. Moreover, such approximation will not change the smoothness property of the restored signal because here the restored signal is still zero if the input signal is zero. The condition Eq. (11) implies that the redundant filters are not necessary for the signal recovery, which is more general than the explanation from Ye *et al.*, wherein the number of filters increases in the decoding phase. In addition, unlike the work by Ye *et al.*, our analysis considers the nonlinearity, which is the key ingredient of deep learning.

**Remark 2:** It can be seen that Soft-AE matches *VisuShrink* and *BayesShrink* more closely than the other aforementioned variants of wavelet shrinkage algorithms. In Soft-AE, the thresholds are assigned to each sub-band differently, without estimating a universal threshold from the noise variance and the number of pixels like in *VisuShrink*. In contrast to *BayesShrink,* Soft-AE demands no statistical estimation, and all the thresholds are learned from data.

**Remark 3:** The interpretability of Soft-AE will not be undermined by the addition of residual connections, if residual connections are symmetrically incorporated. In a residual version of Soft-AE, the features to be learned turn into the residual features, which are still modifiable via wavelet shrinkage. Thus, Eq. (9) still holds for the residual features. In addition, Soft-AE will embrace the merits of residual shortcuts. For example, the employment of residual connections will accommodate the training difficulties in deep models. It was mentioned that feed forward neural networks do not excel in learning the identity mapping [39], and residual connections are able to circumvent the gradient explosion/vanishing problems, thereby facilitating the training of deep networks.

Although interpretability is our major motivation, we also would like to argue that Soft-AE has another important merit: adaptivity. In the era of big data, it is hypothesized that the most comprehensive information is contained in big data, and the best tool to dig them out is deep learning. Given $x \in \mathbf{R}$, the soft thresholding unit is conveniently expressed as

$$\eta_{b<0}(x) = \mathrm{ReLU}(x+b) - \mathrm{ReLU}(-x+b), \qquad (12)$$

where $b < 0$ is a trainable parameter. The Soft-AE can adaptively learn optimal wavelet kernels and thresholds through the training process with big data, which empowers Soft-AE with adaptivity and robustness in contrast to traditional wavelet analysis.

*B. Generalized Linear Unit (GenLU)*

In last subsection, we demonstrate the interpretability of autoencoders using soft thresholding units $\eta_{b<0}(x)$. As aforementioned, the utility of $\eta_{b<0}(x)$ in denoising tasks were theoretically justified in Donoho (1995). By symmetry, our curiosity moves to the other side of the coin, that is, we would like to investigate the resolution enhancing property of the activation function: $\eta_{b>0}(x) = sgn(x)\max\{|x|+b, 0\}$ in a super-resolution model. As a result, we further propose a generalized linear unit (GenLU) and its truncated variant (GeLU) in the autoencoder to make it more general.

*B. 1.* $\eta_{b>0}(x)$

Let us first recall two preliminary results regarding the wavelet expansion and a theorem from [18].

**Wavelet Expansion:** Any function $g \in C[0,1]$ has an expansion:

$$g = \sum_{k}^{2^{j_0}-1} \beta_{j_0,k}\tilde{\phi}_{j_0,k} + \sum_{j\geq j_0}\sum_{k}^{2^{j}-1} \alpha_{j,k}\tilde{\psi}_{j,k}\ , \qquad (13)$$

where $\tilde{\phi}_{j_0,k}$ and $\tilde{\psi}_{j,k}$ are from an orthonormal wavelet basis system, such as the Daubechies system. Let $W$ denote the operator such that $W \circ g$ is a vector of coefficients of countable cardinality.

$$y = W \circ g = [\beta_{j_0,\cdot}, \alpha_{j_0,\cdot}, \alpha_{(j_0+1),\cdot}, \dots, \alpha_{j_1,\cdot}, \dots] \qquad (14)$$

Let $T^n$ denote the truncation operator, $(T^n \circ W) \circ g$ generates a vector with the first $n$ entries of $W \circ g$. To put it simply, $W_n = T^n \circ W$ is an empirical wavelet transform that derives the first $n$ coefficients of the transformation of $g$. We define $y^{(n)} = (T^n \circ W) \circ g = W_n \circ g$. Conversely, the empirical inverse transform is implemented by padding zeros with countable entries before the inverse transform: $g' = (W^{-1} \circ P^n) \circ y^{(n)} = W_n^{-1} \circ y^{(n)}$, where $W_n^{-1} = W^{-1} \circ P^n$.

**Theorem [18]:** Suppose $\mathrm{y}_1^{(\mathrm{n})}$ and $\mathrm{y}_2^{(\mathrm{n})}$ are two vectors subsuming truncated empirical wavelet coefficients by $W$, satisfying that $\mathrm{y}_1^{(\mathrm{n})}$ is elementwise smaller than $\mathrm{y}_2^{(\mathrm{n})}$ in absolute value, i. e., $|\mathrm{y}_1^{(\mathrm{n})}| \leq |\mathrm{y}_2^{(\mathrm{n})}|$, if $g_1' = W_n^{-1} \circ \mathrm{y}_1^{(n)}$ and $g_2' = W_n^{-1} \circ \mathrm{y}_2^{(n)}$, then $\|g_1'\|_{B_{p,q}^s} \leq C(s,p,q)\|g_2'\|_{B_{p,q}^s}$, where $C(s,p,q)$ is a constant and $\|\cdot\|_{B_{p,q}^s}$ is the Besov norm that is the smoothness measure family controlled by $(s, p, q)$. For example, the Besov norm of $f$ incorporates a term: $\int_0^\infty \left|\frac{w_p^2(f^{(l)},t)}{t^\alpha}\right|^q \frac{dt}{t}$, where $w_p^2(f^{(l)},t) = \sup_{|h|\leq t} ||\Delta_h^2 f^{(n)}||_p$, $s = l + \alpha$, and $\Delta_h^2 f^{(l)} = f^{(l)}(x-h) - f^{(l)}(x)$. $f^{(l)}$ is the $l^{th}$ derivative of $f$. The utility of $\Delta_h^2 f^{(l)}$ is to measure the extent of oscillation of $f^{(l)}$. When $l = 0$, the smoothness of $f$ is directly revealed by second-order differences.

Without loss of generality, we ignore the down-sampling effect in the observation and assume that the deblurring by $\eta_{b>0}(x)$ is abstracted as

$$\hat{f}_{\mathrm{HR}} = W_n^{-1} \circ \eta_{b>0} \circ W_n \circ [f_{LR} + \epsilon \cdot z], \qquad (15)$$

where $f_{LR}$ is a blurred low resolution (LR) signal of the same size as that of the expected high resolution (HR) recovered signal $\hat{f}_{\mathrm{HR}}$, $\epsilon \cdot z$ is the noise with $z \sim N(0,1)$, and $\epsilon$ is noise intensity. Then, we have the following Proposition:

**Proposition**: Let $\hat{f}_{\mathrm{HR}}$ and $f_{LR}$ be two functions produced by Eq. (15). There is a universal constant $\pi_n$ with $\pi_n \to 1$ as $n \to \infty$, and constant $C(s,p,q)$ depending on the Besov norm and the wavelet basis $\Psi$ such that

$$\mathrm{Prob}\left\{\|f_{LR}\|_{B_{p,q}^s} \leq C(s,p,q)\left\|\hat{f}_{\mathrm{HR}}\right\|_{B_{p,q}^s}\right\} \geq \pi_n. \quad (16)$$

**Remark 4:** Eq. (16) reveals an important relationship between the degraded low-resolution signal and the high-resolution reconstruction. With the overwhelming likelihood and in a broad family of smoothness measure in terms of the Besov norm, the recovered signal $\hat{f}_{\mathrm{HR}}$ is at least as smooth as that of $f_{LR}$, which is to say that the reconstruction is a resolution-elevating process, because usually the high-resolution signal is less blurred and tend to have higher score in terms of some smoothness metric. What's more, if the authentic signal is zero, then the sampled observed signal should be zero as well. Eq. (16) conforms to such an expectation.

Now, let us analyze the correctness of our proposition. We define

$$y_{LR} + \delta \cdot u_I \equiv W_{\mathrm{n}} \circ [f_{LR} + \epsilon \cdot z], \quad (17)$$

where $y_{LR}$ corresponds to $W_{\mathrm{n}} \circ f_{LR}$, and $\delta \cdot u_I$ corresponds to $W_{\mathrm{n}} \circ (\epsilon \cdot z)$. For now, we presume that $u_I$ is deterministic and ignore its probabilistic character. Then, we define

$$\hat{y}_{\mathrm{HR}} \equiv \eta_{b>0} \circ [y_{LR} + \delta \cdot u_I], \quad (18)$$

where $u_I$ satisfies $|u_I| \leq 1$, $\delta > 0$ denotes intensity, $I_n$ is the index set of cardinality $n$, and $\hat{f}_{\mathrm{HR}} = W_{\mathrm{n}}^{-1} \circ \hat{y}_{\mathrm{HR}}$. By setting $b = \delta$, we obtain $\hat{y}_{HR}^{\delta} = \eta_{b=\delta}(y_{LR} + \delta \cdot u_I)$, then $\hat{y}_{HR}^{\delta}$ is elementwise greater than $y_{LR}$ in the absolute sense. Mathematically,

$$|(\hat{y}_{HR}^{\delta})_I| \geq |(\mathrm{y}_{\mathrm{LR}})_I|, \ \forall I \in I_n \quad (19)$$

The reason is that in each coordinate $I$, $(\hat{y}_{HR}^{\delta})_I$ satisfies

$$|(\hat{y}_{HR}^{\delta})_I| = ||(\mathrm{y}_{\mathrm{LR}})_I + \delta \cdot u_I| + \delta|$$
$$\geq ||(\mathrm{y}_{\mathrm{LR}})_I + \delta \cdot u_I| + \delta|u_I|| \geq |(\mathrm{y}_{\mathrm{LR}})_I| \quad (20)$$

Then, we move back that $u_I$ are actually independently and identically distributed noise. We utilize the following fact regarding a random vector that if $u_I$ is independently and identically distributed with $N(0,1)$, then

$$\mathrm{Prob}\left\{\sup_{I \in I_n}|u_I| \leq \sqrt{2logn}\right\} \to 1, n \to \infty. \quad (21)$$

If we set $b = \delta = \sqrt{2logn}\,\epsilon$, we will arrive at

$$\mathrm{Prob}\left\{|(\hat{y}_{HR}^{\delta})_I| \geq |(\mathrm{y}_{\mathrm{LR}})_I|, \forall I \in I_n\right\} \to 1, n \to \infty. \quad (22)$$

Eq. (22) implies that wavelet coefficients $|(\hat{y}_{HR}^{\delta})_I|$ are very likely to be greater than $|(\mathrm{y}_{\mathrm{LR}})_I|$ for $\forall I \in I_n$. Then, utilizing the aforementioned theorem and noting $\hat{f}_{\mathrm{HR}} = W_{\mathrm{n}}^{-1} \circ \hat{y}_{HR}^{\delta}$ and $f_{LR} = W_n^{-1} \circ \mathrm{y}_{\mathrm{LR}}$, we arrive at

$$\mathrm{Prob}\left\{\|f_{LR}\|_{B_{p,q}^s} \leq C(s,p,q)\left\|\hat{f}_{\mathrm{HR}}\right\|_{B_{p,q}^s}\right\} \geq \pi_n. \quad (23)$$

*B. 2. GenLU and GeLU*

Inspired by the effectiveness of the soft thresholding unit $\eta_{b<0}(x)$ for denoising and the potential resolution enhancement property of $\eta_{b>0}(x)$ implied by the preceding analysis, we are motivated to unify them into a generalized linear unit (GenLU) to empower the autoencoder, and demonstrate its utilities for both denoising and deblurring. The rationale is that each neuron is able to adapt its bias towards either inhibiting noise appearance or enhancing subtle features during the training. The capability unlocked by GenLU can be straightforwardly formulated as

$$\mathrm{GenLU}(x) = sgn(x)\max\{|x| + b, 0\}, \quad (24)$$

where $b$ is an arbitrary real number to be learned. Naturally, we can have GeLU by suppressing the negative part of the input to promote the sparsity. Mathematically, GeLU is expressed as

$$\mathrm{GeLU}(x) = \max\{\mathrm{GenLU}(x), 0\}. \quad (25)$$

Note that ReLU is now embedded in GeLU. The activation patterns of GenLU and GeLU are shown in Figure 4. Coincidently, Hendrycks and Gimpel [50] proposed the Gaussian Error Linear Unit (GELU), a statistically random activation function that is clearly different from our proposed GeLU.

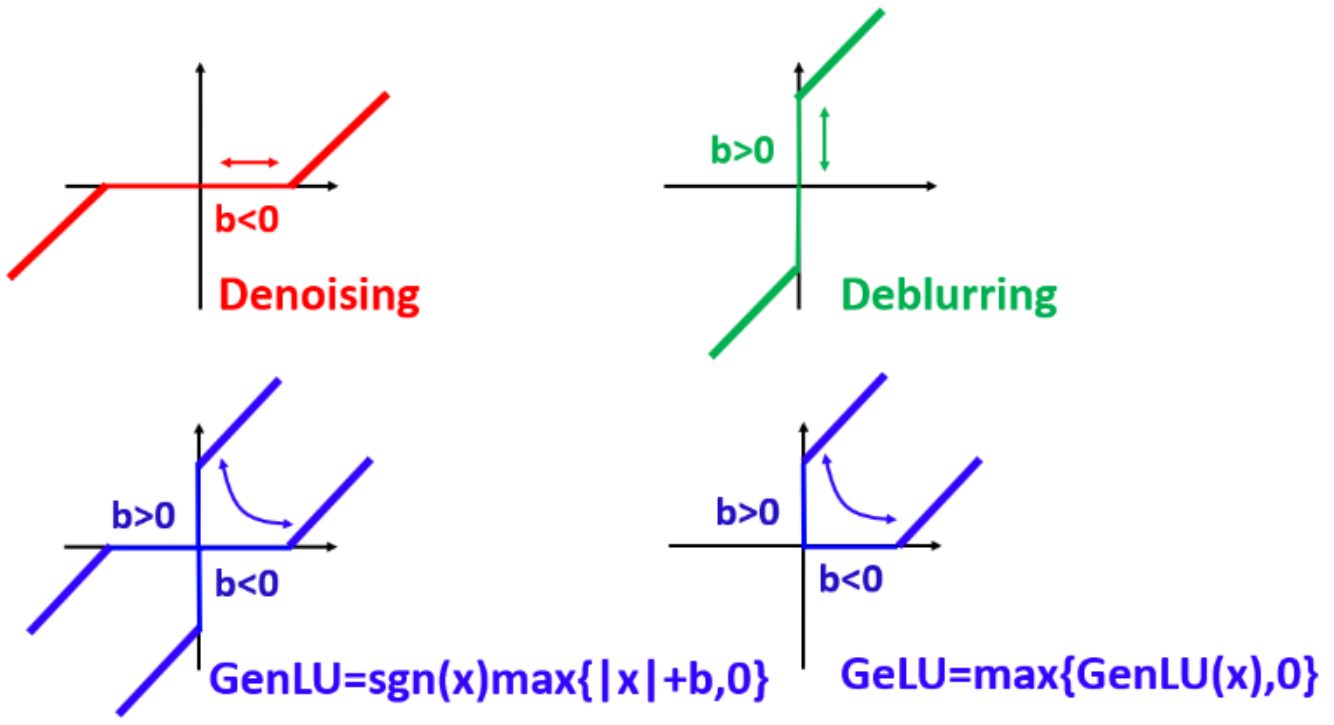

Figure 4. Various filtration functions, where the arrow directions indicate how the activation pattern may change. (a) the soft-thresholding unit for denoising; (b) when $b > 0$, the filtering unit enhances the resolution; (c) the activation pattern of a generalized linear unit (GenLU) embraces denoising and deblurring capabilities by turning $b$ as an adaptive parameter; (d) the activation pattern of the truncated GenLU (GeLU). Note that GeLU=ReLU only when $b < 0$.

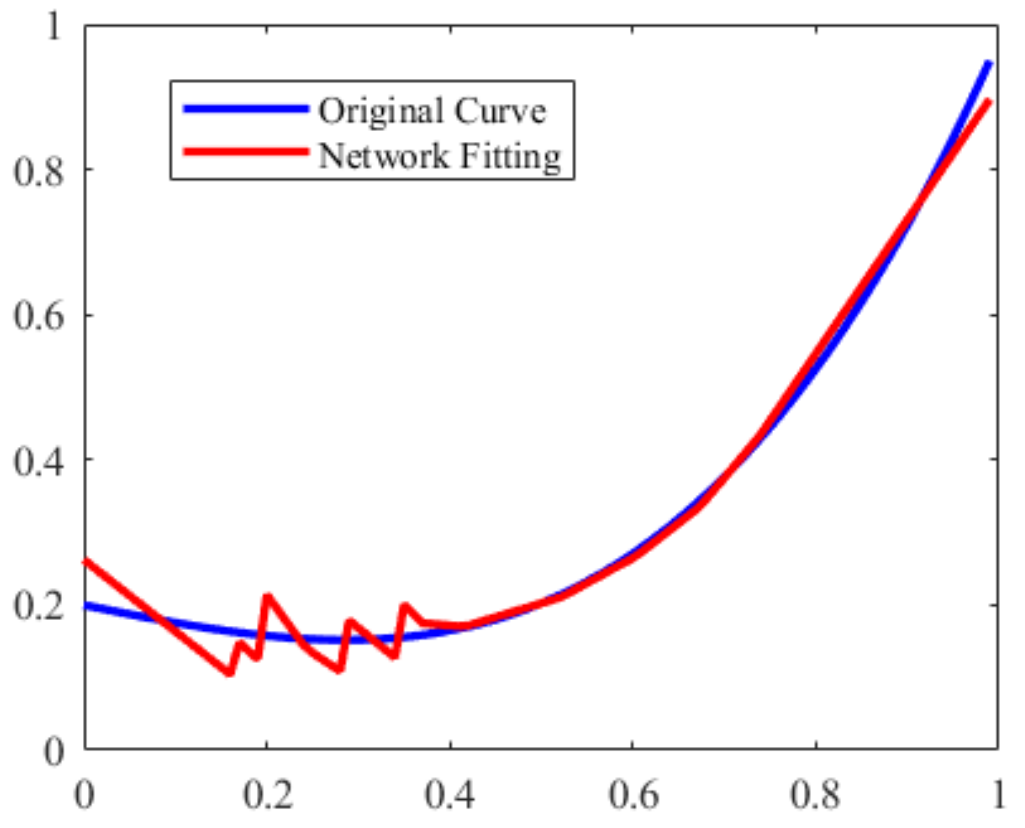

Figure 5. A one-hidden-layer GeLU network is trained to fit the univariate function $f(x) = x^3 - 0.25x + 0.2$ with the synthesized data which are sampled from [0,1] with the interval of 0.01.

Figure 5 shows a toy example where a one-hidden-layer GeLU network is trained to fit the univariate function $f(x) = x^3 - 0.25x + 0.2$ with the synthesized data which are sampled from $[0,1]$ with the interval of 0.01. It is seen that the GeLU network well fits the $f(x)$, particularly in the region of $[0.4,1]$, despite that there are slightly oscillations in the region of $[0, 0.4]$.

## III. EXPERIMENTS

### *A. Denoising Experiments with Soft-AE*

In this section, we compare the performance of our Soft-AE to the state of-the-art networks to justify that Soft-AE not only is interpretable but also performs comparably or favorably in solving real-world problems. Specifically, we selected the convolutional autoencoder with ReLU, Leaky-ReLU, and Concatenated ReLU as the benchmark models. For convenience, we denote them as ReLU-AE, Leaky-AE, and Conc-AE respectively. We can enable a soft thresholding unit with two ReLU units as shown in Eq. (12). Mathematically,

$$\text{LeakyReLU}(x) = \begin{cases} x & if\ x > 0 \\ \alpha x & if\ x < 0 \end{cases} \tag{26}$$

In our TensorFlow environment, $\alpha$ was set to 0.2 by default. Concatenated-ReLU concatenates two ReLU outputs in opposite phases. Mathematically,

$$\text{Concatenate}\ \{\text{ReLU}(x), \text{ReLU}(-x)\} \tag{27}$$

The dimensionality of inputs is doubled after being processed by Concatenated ReLU. Thus, the output of Conc-AE will have an even dimensionality in contrast to those of Soft-AE, Leaky-AE, and ReLU-AE. Because the images in our experiments are of odd channels (either greyscale or RGB images), we use ReLU in the output layer of Conc-AE.

In our experiments, we evaluated the utility of Soft-AE in either the generic or residual structure. For structural fidelity, neither pooling nor un-pooling operations were used. Overall, the loss function for all the models was defined as $L(\Theta) = \frac{1}{N}\sum_i^N ||F(X_i^{noised}; \Theta) - X_i^{label}||^2$, where $\Theta$ denotes hyper-parameters, $X_i^{noised}, X_i^{label}$ are the input and label vectors respectively. Despite the fact that a soft thresholding unit is discontinuous at zero, the empirical results show that this is not an issue in our studies because a unit is still optimizable via gradient-based search using the one-sided derivative which always exists and can be used as needed.

We first tested the denoising performance of different models on natural image benchmarks CIFAR-10 and BSD-300 respectively. CIFAR-10 [40] is a classic benchmark dataset of 50,000 training images and 10,000 test images. Each image is of 32*32 in the RGB format. BSD-300 contains 300 high-quality images of different sizes, where 200 images are for training and 100 images for testing. Because CIFAR-10 is a relatively simple benchmark and BSD-300 is more complicated, we applied the autoencoders of generic structures on CIFAR-10 and autoencoders with residual links on BSD-300, respectively. For further evaluation, we also conducted denoising experiments on the Mayo Clinical Dataset to show that Soft-AE performs well for both natural and medical image denoising tasks. To quantify the denoising performance, we used structural similarity (SSIM) and peak-to-noise ratio (PSNR) as the metrics.

**Denoising on CIFAR-10:** To understand the performance of different models, three typical network structures were evaluated. As shown in TABLE I, they are of (1) four convolutional layers with eight channels in every hidden layer, (2) four convolutional layers with sixteen channels in each hidden layer, and (3) six convolutional layers with sixteen channels per hidden layer. The convolutional kernel size in every layer was set to 3*3. The zero padding was used for convolution to keep the image size intact. In the case of Conc-AE, the activation function for the output layers was configured as ReLU. For symmetry, we used ReLU in the first layer as well. Concatenated ReLU activations were employed for the rest layers. All the activations in ReLU-AE and Leaky-AE were done by ReLU and Leaky-ReLU respectively. In Soft-AE, the encoding part takes soft thresholding units while the decoding part uses linear functions.

TABLE I: THREE CONVOLUTIONAL AUTOENCODER ARCHITECTURES TESTED ON CIFAR-10.

| Architecture | Convolutional Layer | Channel Number | Shortcut |
|---|---|---|---|
| Structure -1 | 4 | 8 | No |
| Structure -2 | 4 | 16 | No |
| Structure -3 | 6 | 16 | No |

All the images were normalized by division with 255. Noisy images were synthesized by adding additive Gaussian noise with zero mean and standard deviation $\sigma = 0.1, 0.15, 0.2$ respectively. Negative pixel values were truncated to 0. In the training process, noisy images were fed into the network, and denoised images were compared with the clean counterparts. With random initializations, each network was trained five times to produce mean SSIM and PSNR values. For all the models, we used the Adam for training. A batch of 50 training samples were processed in every iteration, the number of epochs was set to 20, and the learning rate was $10^{-3}$. The results are summarized in TABLE II. Superscripts 1-3 correspond to the three architectures in TABLE I respectively. The best performance among the four models with respect to the specific noise level is bold-faced. Generally speaking, the four autoencoders share the same trend that the performance goes down as the noise level goes up; all the models of structure-2 and structure-3 yield higher PSNR and SSIM scores than their counterparts of structure-1. It is underlined that Soft-AE keeps the clearly superior performance in a majority of cases, particularly for structure-1. In those cases when Soft-AE does not give the best metrics, it follows the best performer closely. Overall, it is concluded that Soft-AE indeed produces comparable or favorable denoising performance relative to the state-of-the-arts.

**Denoising on BSD-300:** We randomly selected 30,000 patches of 50*50 from the BSD images to single out 20,000 batches for

TABLE II: DENOISING PERFORMANCE COMPARISON AMONG LEAKY-AE, CONC-AE, RELU-AE AND SOFT-AE ON CIFAR-10.

| Metric | $\sigma$ | Leaky-AE$^1$ | Conc-AE$^1$ | ReLU-AE$^1$ | Soft-AE$^1$ | Leaky-AE$^2$ | Conc-AE$^2$ | ReLU-AE$^2$ | Soft-AE$^2$ | Leaky-AE$^3$ | Conc-AE$^3$ | ReLU-AE$^3$ | Soft-AE$^3$ |
|---|---|---|---|---|---|---|---|---|---|---|---|---|---|
| PSNR | 0.1 | 27.043 | 26.961 | 27.150 | **27.469** | 27.936 | 27.640 | 27.919 | **27.944** | 27.898 | 27.815 | 27.974 | **28.039** |
| | 0.15 | 25.058 | 24.957 | 25.186 | **25.370** | 25.752 | 25.676 | 25.783 | **25.786** | 25.914 | 25.837 | **26.036** | 25.774 |
| | 0.2 | 23.845 | 23.606 | 23.913 | **23.952** | 24.393 | 24.320 | **24.403** | 24.355 | **24.572** | 24.385 | 24.537 | 25.535 |
| SSIM(%) | 0.1 | 91.662 | 91.533 | 91.974 | **92.368** | **93.298** | 93.023 | 93.107 | 93.251 | 93.325 | 93.160 | **93.505** | 93.459 |
| | 0.15 | 87.757 | 87.396 | 88.124 | **88.513** | 89.502 | 89.300 | 89.090 | **89.570** | 89.924 | 89.681 | **90.185** | 89.897 |
| | 0.2 | 84.605 | 83.796 | **84.816** | 84.744 | 86.079 | 85.922 | **86.146** | 86.089 | **86.730** | 86.326 | 86.695 | 86.526 |

Note: superscripts 1-3 correspond to three architectures shown in TABLE I.

training, and the remaining for testing. Similarly, we utilized the networks of three symmetric structures to perform comparisons, as shown in TABLE III. Specifically, these networks are in the following structures: (1) eight convolutional layers with 8 channels in each layer, (2) eight convolutional layers with 12 channels in each layer, and (3) ten convolutional layers with 8 channels in each layer. As far as the topologies of skip-connections are concerned, not all paired encoder/decoder layers were bridged by shortcuts for a reasonable computational overhead.

TABLE III: THREE CONVOLUTIONAL AUTOENCODERS USING SKIP CONNECTIONS TESTED ON BSD-300.

| Architecture | Convolutional Layer | Channel Number | Shortcut Topology |
|---|---|---|---|
| Structure -1 | 8 | 8 | 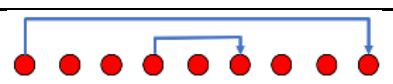 |
| Structure -2 | 8 | 12 | 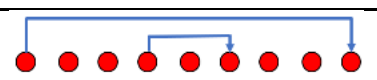 |
| Structure -3 | 10 | 8 | 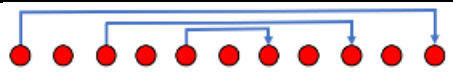 |

All the images were normalized by division with 255. According to the protocols used for CIFAR-10, we synthesized noisy images by adding additive Gaussian noise with zero mean. The standard deviations were set to $\sigma = 0.1, 0.15, 0.2$ respectively. Negative pixel values were lower bounded to 0. Because the initialization was random, each network was trained five times to compute mean SSIM and PSNR values. For all the models ReLU-AE, Leaky-AE, Conc-AE and Soft-AE, we used the Adam for optimization. A batch of 50 training samples were processed per iteration, the number of epochs was 20, and the learning rate was $10^{-3}$.

The denoising results are in TABLE IV. The best performance among the four models for each noise level is bold-faced. With residual connections, Soft-AE performs even better. In the networks of structure-1 and structure-3, Soft-AE performs the best in terms of both SSIM and PSNR for all the noise levels. Particularly, the SSIM and PSNR improvements by Soft-AE are significantly over Conc-AE and ReLU-AE. However, the counterexamples exist for Soft-AE$^2$, since the best performances in some cases are from Leaky-AE$^2$, but the PSNR and SSIM values achieved by Soft-AE$^2$ are very close to those of Leaky-AE$^2$. To check if the superiority is really significant or not, we further conducted statistical hypothesis testing and the results were put into the supplementary material for more information.

**Denoising on Low-dose CT:** Low-dose CT imaging has gained a considerable traction over the past decade due to its potential to reduce the risk induced by X-ray radiation to a patient. One effective way to reduce the X-ray dose is to use a lower X-ray flux. However, a reduced X-ray flux will elevate image noise and compromise image quality. Currently, the algorithms dedicated to low-dose CT image denoising can be roughly put into three categories: (a) sinogram domain filtering, (b) iterative reconstruction, and (c) image post-processing. The sinogram filtering methods [41-43] can be used when the data format and noise characteristics are known. Nevertheless, sinogram filtering tends to reduce spatial resolution. On the other hand, the image-domain iterative methods were extensively investigated, especially model-based and compressed sensing methods [44-46]. Although modern iterative algorithms produce encouraging results, their computational cost is rather high. Finally, the image post-processing methods, such as block matching [47-48], are directly applied to low-dose CT images without any direct access to raw data. The main barrier for the post-processing methods is that the noise distribution cannot be perfectly pre-determined, leading to structural blurring or distortion.

Recently, deep learning methods were successfully applied to low-dose CT denoising, such as RED-CNN [49], which delivered a competitive denoising performance. Here we tested

TABLE IV: DENOISING PERFORMANCE COMPARISON AMONG LEAKY-AE, CONC-AE, RELU-AE AND SOFT-AE ON BSD-300

| Metric | $\sigma$ | Leaky-AE$^1$ | Conc-AE$^1$ | ReLU-AE$^1$ | Soft-AE$^1$ | Leaky-AE$^2$ | Conc-AE$^2$ | ReLU-AE$^2$ | Soft-AE$^2$ | Leaky-AE$^3$ | Conc-AE$^3$ | ReLU-AE$^3$ | Soft-AE$^3$ |
|---|---|---|---|---|---|---|---|---|---|---|---|---|---|
| PSNR | 0.1 | 29.252 | 28.507 | 28.789 | **29.543** | **29.437** | 29.367 | 29.336 | 29.363 | 29.545 | 29.037 | 29.425 | **29.700** |
| | 0.15 | 26.999 | 26.470 | 26.786 | **27.486** | 27.424 | 27.121 | 27.287 | **27.432** | 27.349 | 26.875 | 27.275 | **28.109** |
| | 0.2 | 25.589 | 24.462 | 25.406 | **26.153** | 26.064 | 25.829 | 25.955 | **26.094** | 25.797 | 25.237 | 25.739 | **26.267** |
| SSIM(%) | 0.1 | 89.803 | 88.892 | 88.803 | **90.227** | 90.160 | **90.181** | 89.699 | 89.916 | 90.511 | 89.932 | 90.212 | **90.584** |
| | 0.15 | 84.372 | 83.081 | 83.566 | **85.315** | **85.381** | 84.727 | 85.089 | 85.186 | 85.298 | 84.196 | 85.124 | **85.949** |
| | 0.2 | 79.504 | 78.485 | 78.718 | **80.997** | **81.129** | 80.596 | 80.823 | 80.876 | 80.208 | 79.042 | 80.139 | **81.450** |

Note that superscripts 1-3 correspond to the three architectures in TABLE III respectively.

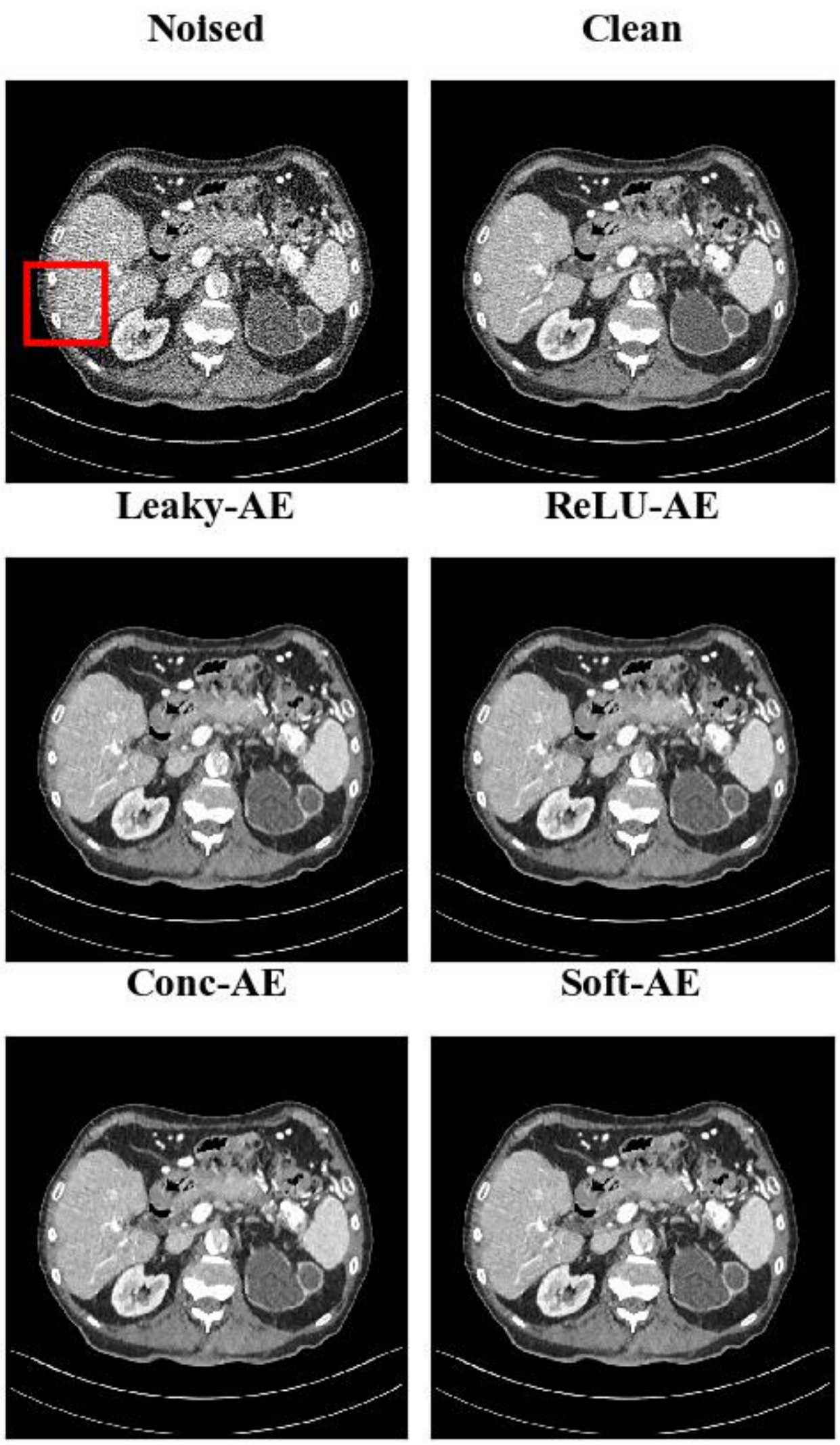

Figure 6. The comparison of denoising results from different models for an abdominal region. Display window is [-240,160].

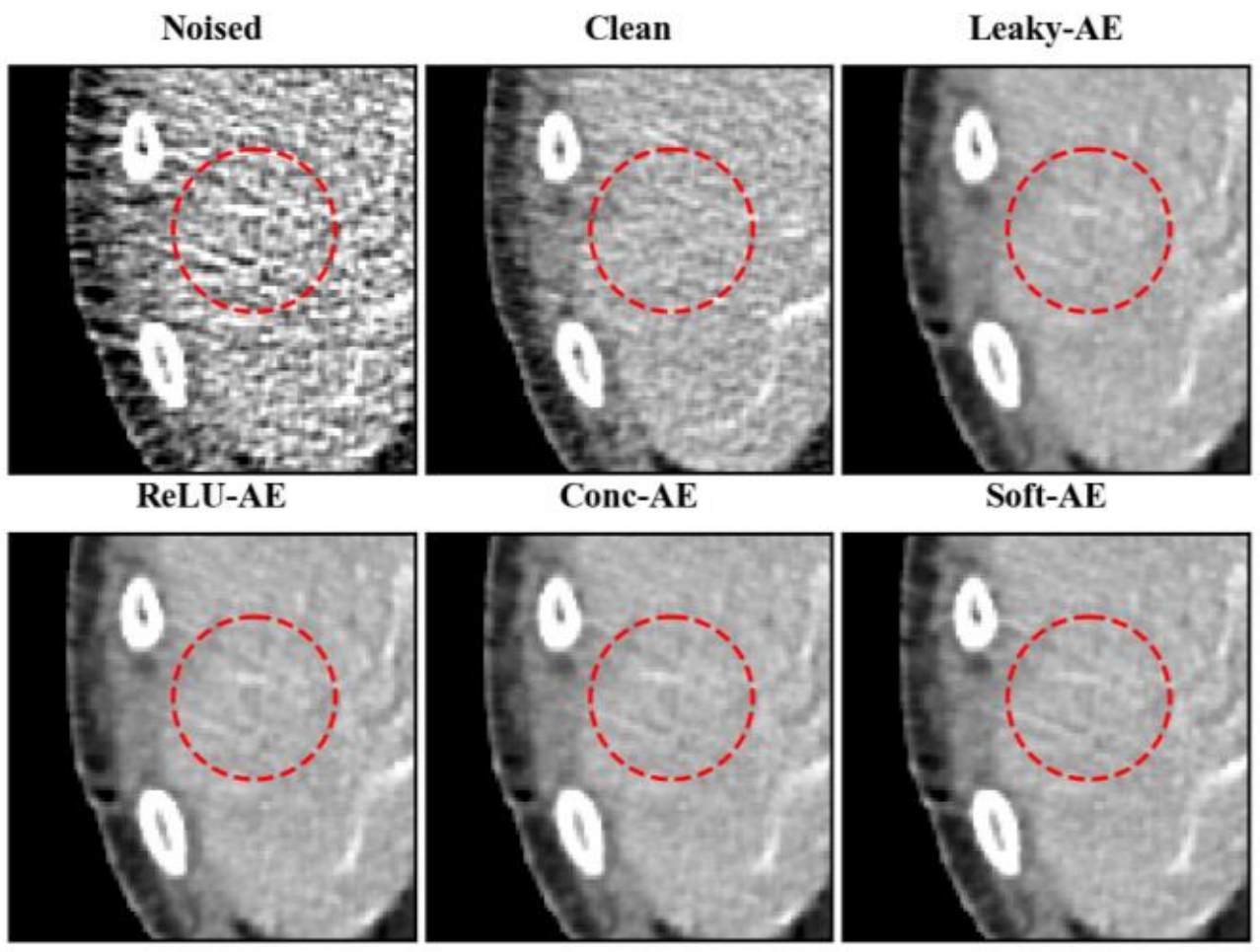

Figure 7. Zoomed regions from Figure 6. The red circle highlights the region in high-contrast as best revealed by Soft-AE. Display window is [-240,160].

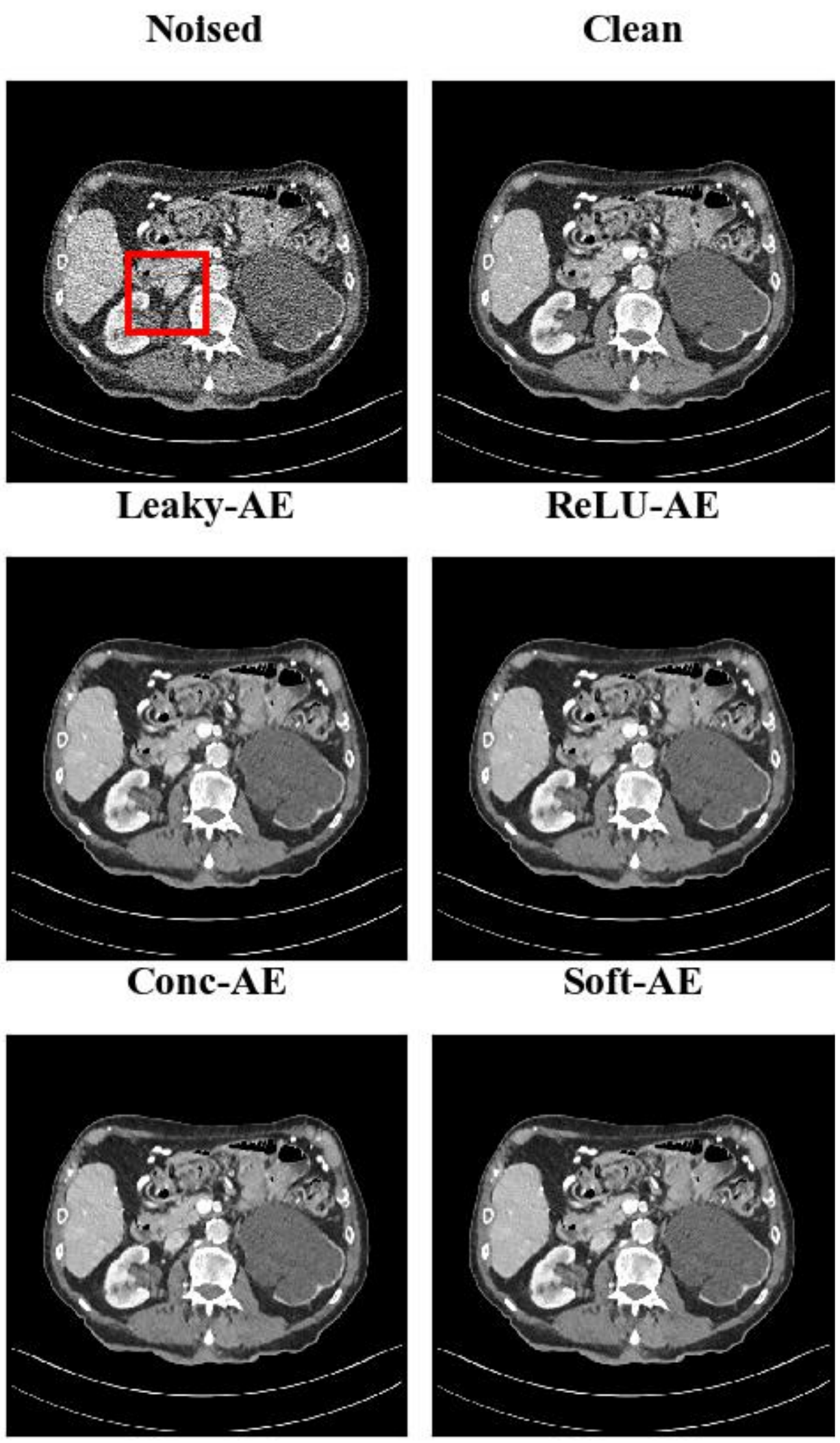

Figure 8. The comparison of denoising results from different models for an abdominal region. Display window is [-240,160].

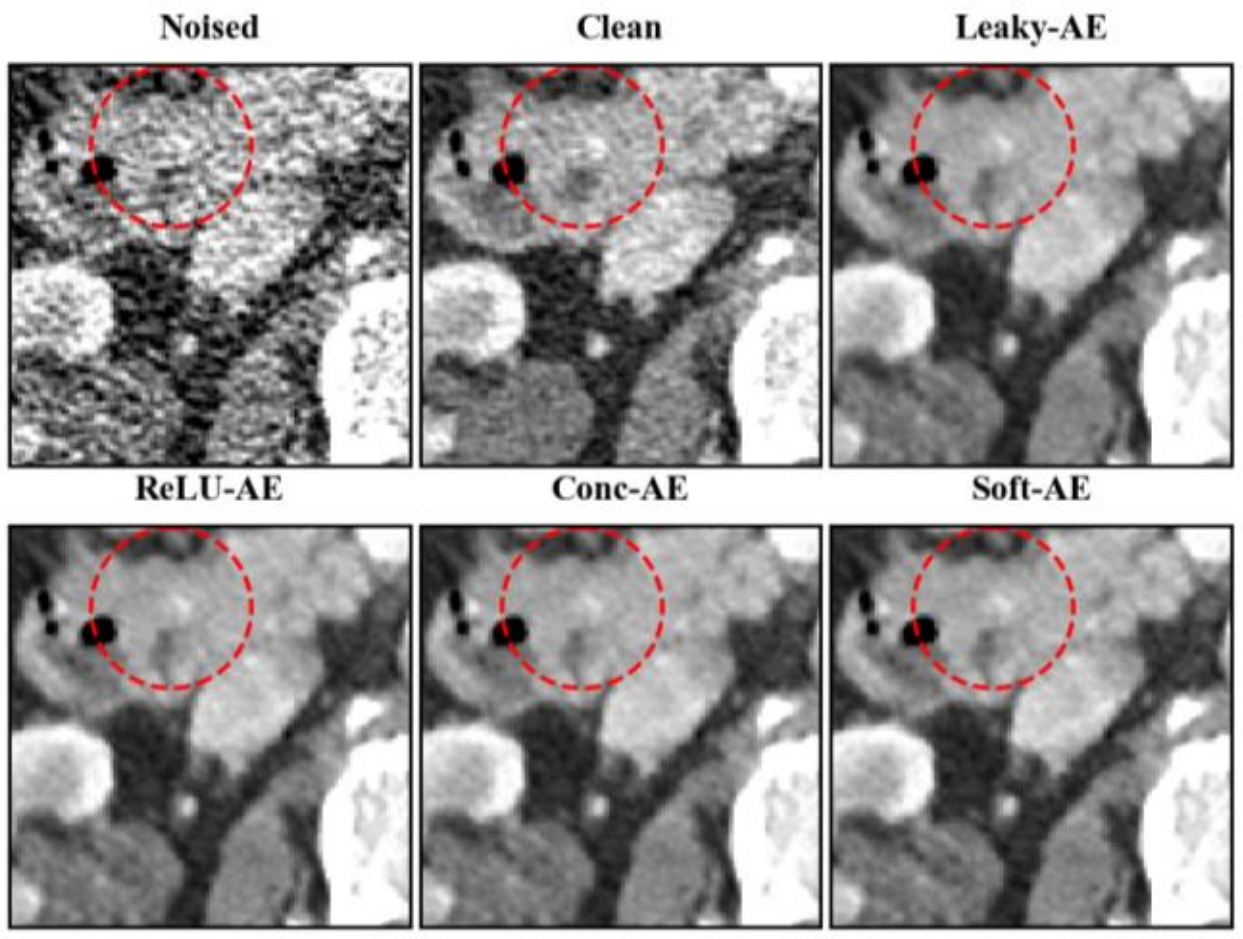

Figure 9. Zoomed regions from Figure 8. The red circle highlights the region in high-contrast as best revealed by Soft-AE. Display window is [-240,160].

the performance of our Soft-AE for low-dose CT denoising with a real clinical dataset prepared by Mayo Clinics for "the 2016 NIH-AAPM-Mayo Clinic Low Dose CT Grand Challenge" [51]. This dataset has 2,378 full dose and corresponding quarter dose 512*512 CT images from 10 patients. In our study, we randomly extracted 64,000 64*64 patches from these images for training. After the training, we tested the models based on independent full-size images. For CT denoising, we employed ReLU-AE, Leaky-AE and Conc-AE of residual connections. The networks of structure-2 were utilized. In Soft-AE, we used 34 layers with 8 convolutional kernels in each layer. The hyper-parameters for training include 50 batches in each iteration, the learning rate for Adam optimization $1.5 \times 10^{-3}$ in the first 20 epochs, and $1.0 \times 10^{-3}$ in the final 10 epochs.

TABLE V: QUANTITATIVE COMPARISON BETWEEN AE USING QUADRATIC ACTIVATION AND QUADRATIC AUTOENCODER.

| | Fig. 6 | | | Fig. 8 | | |
|---|---|---|---|---|---|---|
| | SSIM | PSNR | RMSE | SSIM | PSNR | RMSE |
| Noised | 0.8131 | 23.502 | 0.06681 | 0.8491 | 25.239 | 0.05471 |
| Leaky-AE | 0.8639 | 28.685 | 0.03679 | 0.9180 | 30.183 | 0.03096 |
| ReLU-AE | 0.8945 | 28.729 | 0.03660 | 0.9185 | 30.231 | 0.03079 |
| Conc-AE | 0.8943 | 28.717 | 0.03665 | 0.9183 | 30.199 | 0.03091 |
| Soft-AE | **0.8951** | **28.731** | **0.03657** | **0.9189** | **30.244** | **0.03074** |

Two representative abdominal CT slices (the 100th and 130th slices from patient L506) were selected to evaluate the performance of Soft-AE and the other models, as shown in Figure 6-9. For better visualization, we zoomed the regions of interest (ROIs) marked by the red rectangles. It is noted that all the models demonstrate denoising effects with different image quality impressions. Figure 7 highlights high structural fidelity by Soft-AE. The structures processed by Leaky-AE, ReLU-AE, and Conc-AE are disappearing. Figure 8 showcases that the results of Leaky-AE, Conc-AE, and ReLU-AE blur some structural details.

TABLE V lists the quantitative comparative results for these images. In both Figures 6 and 8, the highest PSNR and SSIM values and lowest root mean square errors (RMSEs) are from Soft-AE. It is concluded that Soft-AE can deliver a competitive performance compared with its counterparts in this real-world benchmark.

To test the utility and robustness of Soft-AE, we also compared Soft-AE with the *VisuShrink* and *BayesShrink* algorithms and discussed the important parameters in Soft-AE. The results are in the supplementary materials.

### *B. Deblurring and Denoising Experiments with GenLU*

We tested the effectiveness of GenLU for image denoising and deblurring in comparison with Leaky-ReLU and ReLU based on SRCNN [52] that is a forerunner in deep learning for super-resolution. For convenience, we denote SRCNN with ReLU, Leaky-ReLU, and GenLU as ReLU-SRCNN, Leaky-SRCNN, and GenLU-SRCNN respectively. Note that GenLU-SRCNN is more interpretable, since we already have a theoretical basis with $\eta_{b>0}(x)$ for signal recovery. Practically, the employment of GenLU strengthens the flexibility of the network and facilitates its representation. From the perspective of network modularity and functional decomposition, it makes sense that neurons with $\eta_{b>0}(x)$ enhance resolution, while neurons with $\eta_{b<0}(x)$ suppress noise. Thus, the utility of each neuron is indicated by the sign of $b$ that can be learned.

TABLE VI: BASIC PARAMETERS OF SRCNN

| | Layer 1 | Layer 2 | Layer 3 |
|---|---|---|---|
| SRCNN | 64 $9 \times 9$ filters | 32 $1\times 1$ filters | 32 $5\times 5$ filters |

The key features of the SRCNN model are summarized in TABLE VI. We used the same training images as that in [52-53], where the training set consists of 91 images. The Set14 dataset of 14 images was used to evaluate the performance at a magnification factor 3. The preparation of training images followed the same protocol as that in [52]. The low-resolution noisy images were obtained by blurring the original images (sub-sampling by a factor 3 and upscaling it with the same factor) and adding the Gaussian noise with mean 0 and deviation 0.005. Roughly, 24,800 pairs of high-resolution and low-resolution images of $32 \times 32$ were used for training. In our experience, such a training set was sufficiently large to train all the three networks.

TABLE VII: THE PSNR AND RMSE VALUES ON THE SET 14 DATASET.

| Set14 Image | ReLU-SRCNN | | Leaky-SRCNN | | GenLU-SRCNN | |
|---|---|---|---|---|---|---|
| | PSNR | RMSE | PSNR | RMSE | PSNR | RMSE |
| Baboon | 19.006 | 0.01257 | **19.011** | **0.01252** | 18.863 | 0.01299 |
| Barbara | 22.991 | 0.00502 | 22.986 | 0.00506 | **23.159** | **0.00483** |
| Bridge | 21.639 | 0.00686 | 21.613 | 0.00690 | **22.132** | **0.00612** |
| Coastguard | 24.045 | 0.00393 | 24.038 | 0.00395 | **24.276** | **0.00373** |
| Comic | **19.602** | **0.01096** | 19.575 | 0.01102 | 19.477 | 0.01127 |
| Face | 28.697 | 0.00135 | 28.686 | 0.00135 | **28.788** | **0.00132** |
| Flowers | 24.222 | 0.00378 | 24.204 | 0.00380 | **24.400** | **0.00363** |
| Foreman | 26.789 | 0.00209 | 26.766 | 0.00211 | **26.815** | **0.00208** |
| Lenna | 24.335 | 0.00368 | 24.276 | 0.00374 | **24.828** | **0.00329** |
| Mam | 23.623 | 0.00434 | **23.631** | **0.00433** | 22.767 | 0.00529 |
| Monarch | 24.147 | 0.00385 | 24.188 | 0.00381 | **25.109** | **0.00308** |
| Pepper | 28.572 | 0.00139 | 28.517 | 0.00141 | **29.091** | **0.00123** |
| PPT3 | 15.253 | 0.00298 | 15.241 | 0.02992 | **16.745** | **0.00212** |
| Zebra | **22.422** | **0.00572** | 22.416 | 0.00573 | 22.091 | 0.00618 |

For all the three networks, the loss function was MSE. We performed training with a batch size of 128. To facilitate convergence, we initialized networks with a pre-trained model publicly available in GitHub [54], which is an independent implementation of SRCNN on TensorFlow. The learning rate was set to $10^{-4}$. The total number of iterations was 15000.

PSNR and MSE for different models were computed, as summarized in TABLE VII. For each image, the best scores are bold-faced. Overall, the results of ReLU-SRCNN and Leaky-SRCNN on different images are quite close to each other. The highlight is that the proposed GenLU-SRCNN produces the highest PSNR and the lowest MSE values in 10 images among all the 14 images in Set14. Importantly, on the images "Lenna", "Monarch", and "PPT3", GenLU-SRCNN outperforms its counterparts by a large margin ($\geq$ 0.5dB). Figures 10 and 11 show the recovered results, ROIs, and profiles for the compared models on the four images. It can be observed in Figure 11 that all the three networks resolve meaningful details from the degraded images. Furthermore, GenLU-SRCNN produces the most desirable outcomes over the other models. As indicated by the green arrows, the GenLU-SRCNN images preserve sharp edges which are faithful to the labels.

## IV. CONCLUSION

In conclusion, we have investigated to replace the ReLU activation in the setting of convolutional autoencoders and introduced a pair of ReLU units emulating soft thresholding, thereby offering the network interpretability while enhancing the network performance through adaptivity as well. As a result, we have interpreted our Soft-AE as a deeply learned nonlinear wavelet shrinkage system. Furthermore, we have proposed GenLU for more image processing tasks. Interestingly, the function decomposability between different neurons are realized by the tuning of thresholds. In the future, other low-level computer vision tasks such as image impainting can be revisited in our framework.

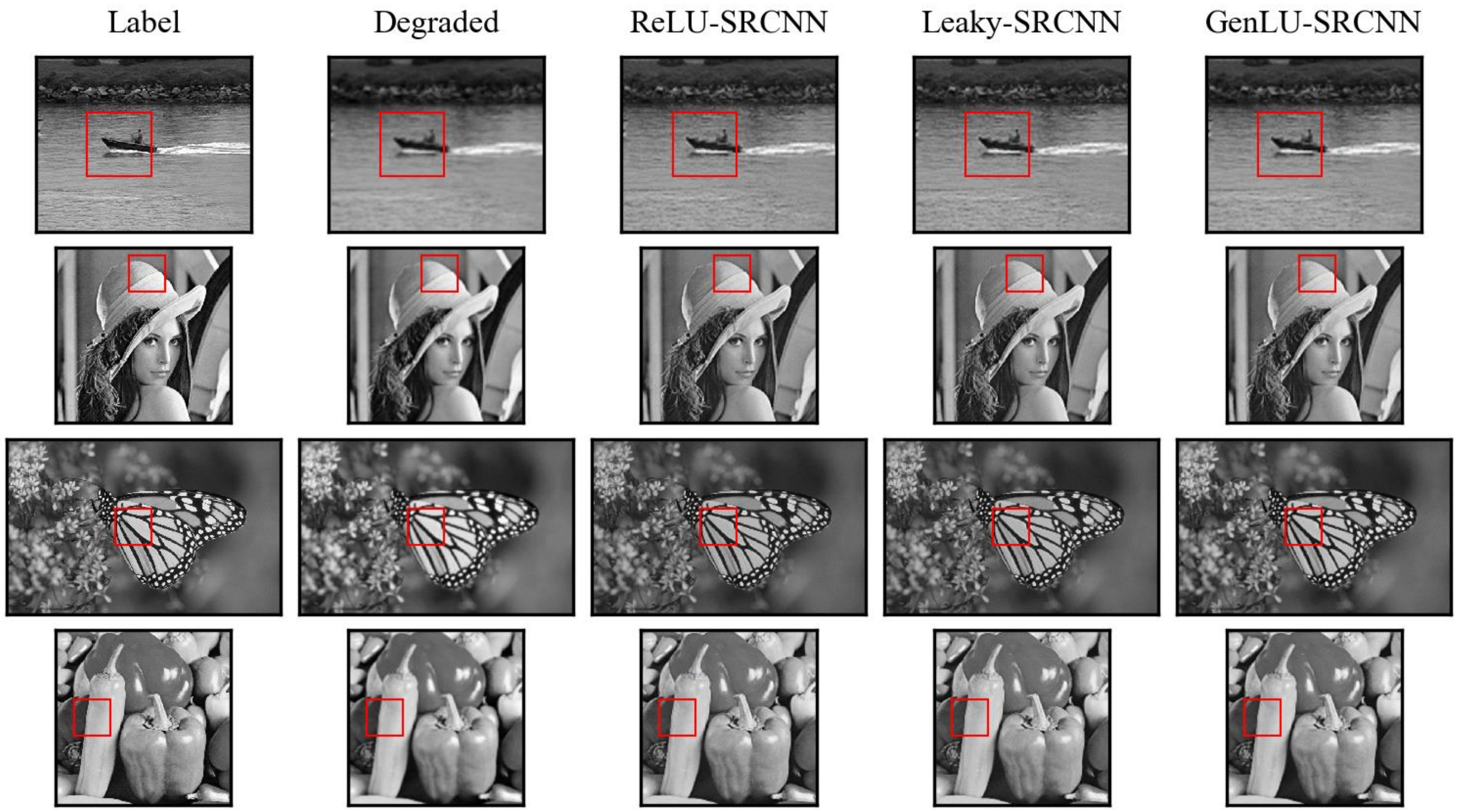

Figure 10. Comparison of recovered images from different models.

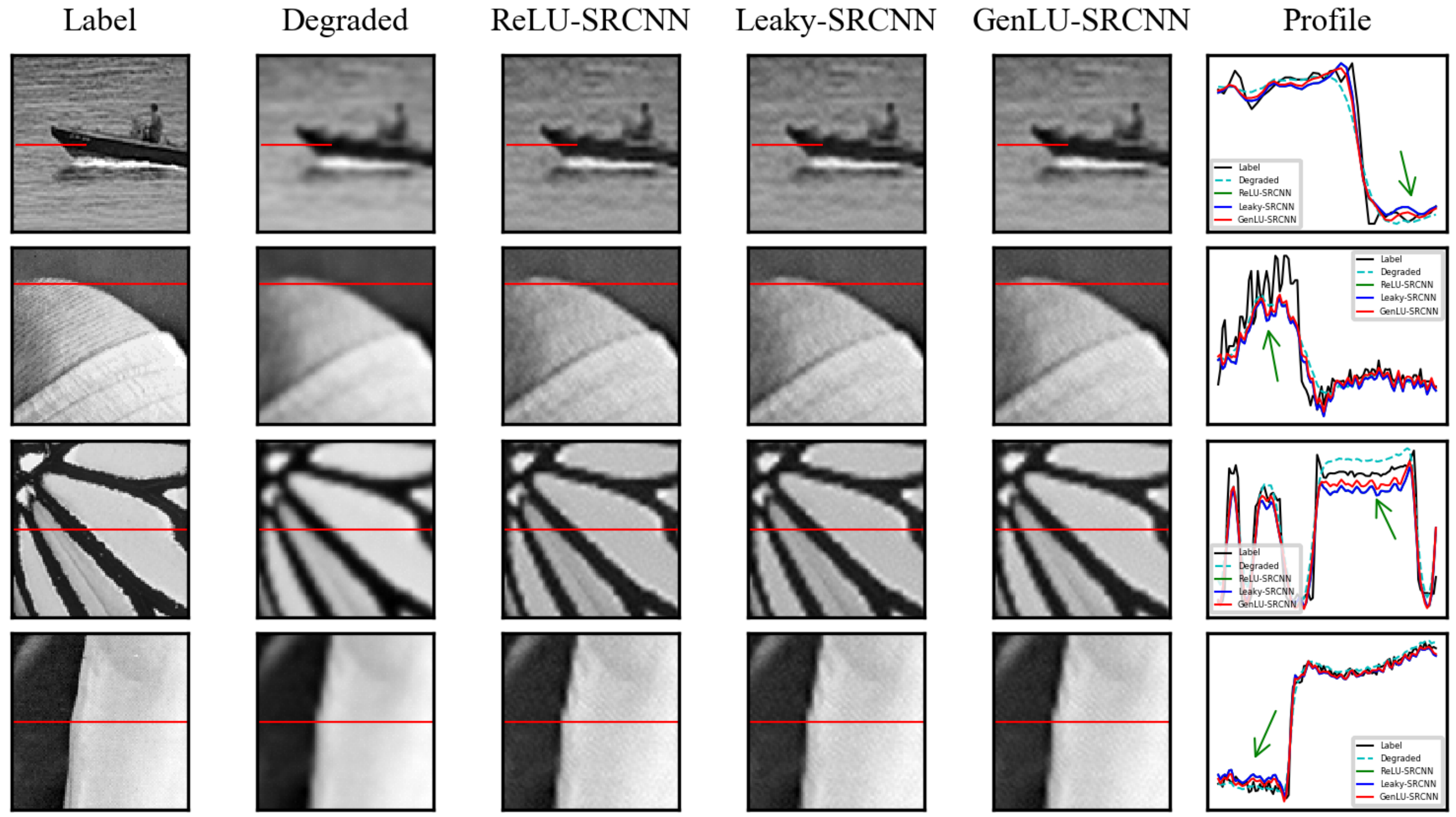

Figure 11. Comparison of ROIs of recovered images and profiles from different models.